\documentclass[journal]{IEEEtran}
\usepackage{amsmath,amsfonts}
\usepackage{algorithmic}
\usepackage{algorithm}
\usepackage{array}

\usepackage[caption=false,font=footnotesize]{subfig}
\usepackage{textcomp}
\usepackage{stfloats}

\usepackage{verbatim}
\usepackage{graphicx}
\usepackage{cite}
\usepackage{multirow}

\usepackage{tabularx}
\usepackage{booktabs}
\usepackage{enumitem}

\usepackage{times}
\usepackage{helvet}
\usepackage{courier}
\usepackage{xcolor}

\usepackage{url}
\usepackage{hyperref}

\usepackage{graphicx}
\usepackage{amsmath}
\usepackage{array}
\usepackage{hhline}

\usepackage{bm}

\usepackage{booktabs,tabularx,pifont}
\usepackage{placeins}
\usepackage{orcidlink}

\begin{document}

\title{DPTrack: Directional Kernel-Guided Prompt Learning
	for \\ Robust Nighttime Aerial Tracking}

\author{Zhiqiang Zhu,
	Xinbo Gao,~\IEEEmembership{Fellow,~IEEE,}
	Wen Lu,~\IEEEmembership{Member,~IEEE,}
	Jie Li,
	Zhaoyang Wang,
	Mingqian Ge
   
\thanks{This work was supported in part by the National Natural Science  Foundation of China under Grants No. 62036007 and U22A2096; in part by the National Natural Science Foundation of China (No. 62476207), the Chongqing Natural Science Foundation Innovation and Development Joint Fund Project under Grant CSTB2023NSCQ-LZX0085. (Corresponding authors: Xinbo Gao.)

Xinbo Gao is with the School of Electronic Engineering, Xidian University, Xi’an 710071, China (e-mail: \href{xbgao@mail.xidian.edu.cn}{\nolinkurl{xbgao@mail.xidian.edu.cn}}).

Zhiqiang Zhu, Wen Lu, Jie Li, Zhaoyang Wang, and Mingqian Ge are with the Visual Information Processing Laboratory, School of Electronic Engineering,  Xidian University, Xi’an, Shaanxi 710071, China (e-mail: \href{mailto:zhuzhiqiang@stu.xidian.edu.cn}{\nolinkurl{zhuzhiqiang@stu.xidian.edu.cn}};
\href{mailto:luwen@xidian.edu.cn}{\nolinkurl{luwen@xidian.edu.cn}}; \href{mailto:leejie@mail.xidian.edu.cn}{\nolinkurl{leejie@mail.xidian.edu.cn}}; \href{mailto:zywang23@stu.xidian.edu.cn}{\nolinkurl{zywang23@stu.xidian.edu.cn}}; \href{mailto:mqge@stu.xidian.edu.cn}{\nolinkurl{mqge@stu.xidian.edu.cn}}).
	}
}

\markboth{Journal of \LaTeX\ Class Files,~Vol.~14, No.~8, August~2021}%
{Shell \MakeLowercase{\textit{et al.}}: A Sample Article Using IEEEtran.cls for IEEE Journals}



\maketitle

\begin{abstract}
Existing nighttime aerial trackers based on prompt learning rely solely on spatial localization supervision, which fails to provide fine-grained cues that point to target features and inevitably produces vague prompts. This limitation impairs the tracker’s ability to accurately focus on the object features and results in trackers still performing poorly. To address this issue, we propose DPTrack, a prompt-based aerial tracker designed for nighttime scenarios by encoding the given object’s attribute features into the directional kernel enriched with fine-grained cues to generate precise prompts. Specifically, drawing inspiration from visual bionics, DPTrack first hierarchically captures the object’s topological structure, leveraging topological attributes to enrich the feature representation. Subsequently, an encoder condenses these topology-aware features into the directional kernel, which serves as the core guidance signal that explicitly encapsulates the object’s fine-grained attribute cues. Finally, a kernel-guided prompt module built on channel–category correspondence attributes propagates the kernel across the features of the search region to pinpoint the positions of target features and convert them into precise prompts, integrating spatial gating for robust nighttime tracking. Extensive evaluations on established benchmarks demonstrate DPTrack's superior performance. Our code will be available at \url{https://github.com/zzq-vipsl/DPTrack}.
\end{abstract}


\section{Introduction}
\IEEEPARstart{V}{isual} object tracking plays an indispensable role in aerial imagery applications, including navigation, trajectory planning, and remote sensing. Given an object’s initial position in the first aerial frame, visual object tracking aims to continuously estimate its position and scale throughout the video \cite{r:yin,r:wu,r:li}. Recently, similarity matching-based trackers have become the mainstream, which learn a similarity network on large-scale datasets to locate the object by matching the feature template with the search region.
\begin{figure}[t]
	\centering
	\subfloat[Pipeline of existing prompt-based trackers.]
	{\includegraphics[width=1\linewidth]{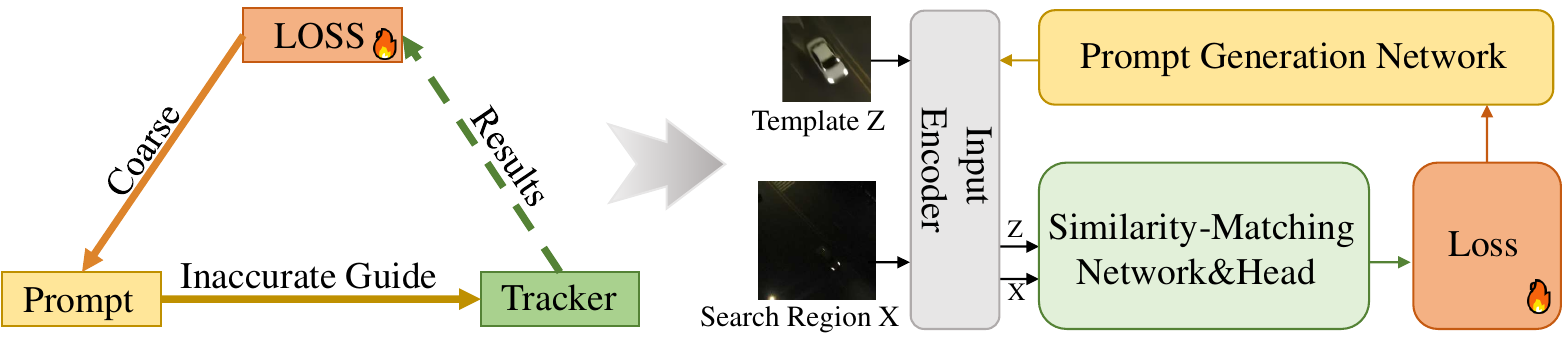}%
		\label{fig:difference_a}}\hfill
	\subfloat[Pipeline of the proposed DPTrack]
	{\includegraphics[width=1\linewidth]{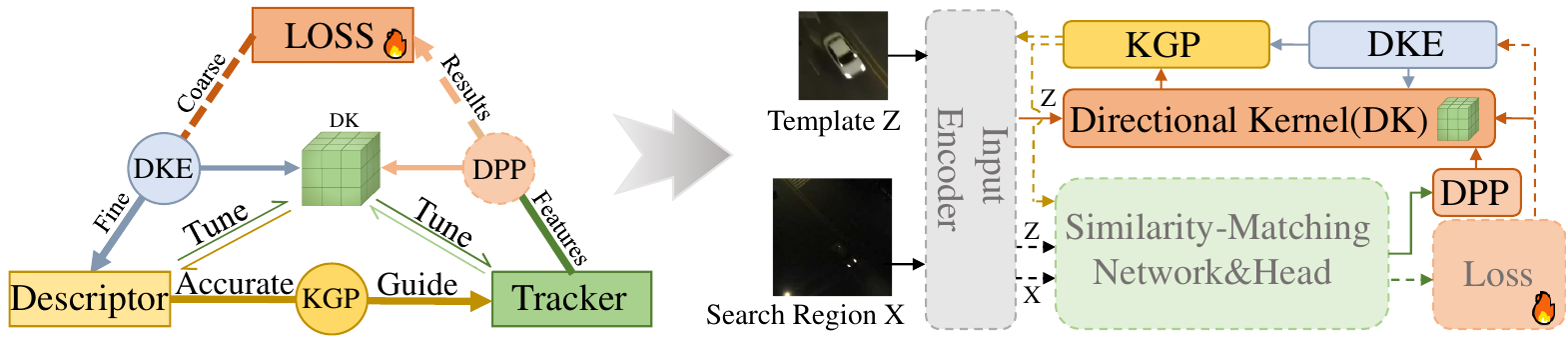}%
		\label{fig:difference_b}
	}
	\caption{Comparison between existing prompt-based trackers and DPTrack. The left illustrates the design philosophy, and the right shows the pipeline. (a): Existing trackers generating prompts solely rely on loss. (b): DPTrack utilizes fine-grained guidance signals to produce accurate prompts.}
	\label{fig:difference}%
\end{figure}

 Although aerial tracking has made notable progress, most existing trackers are designed for ideal lighting conditions, so that they fail to effectively perceive objects obscured in darkness under low-light scenarios (e.g., nighttime)\cite{c:guo2020zero}, 
 thereby undermining the similarity matching mechanism that relies on clear features. Some trackers attempt to address this issue via low-light enhancement \cite{c:fu2022highlightnet,c:li2021adtrack,c:ye2021darklighter,c:zhang2025mambatrack} or domain adaptation \cite{c:zhang2023progressive,c:ye2022unsupervised,c:zuo2024dadiff,c:wu2025lvptrack}, but the introduction of exogenous noise, such as artifacts or false labels, still limits practical performance. 

In recent works, prompt learning has emerged as a promising solution by generating prompts that embed prior knowledge to guide daytime-trained trackers in perceiving the object while avoiding exogenous noise. \textbf{However, due to these trackers relying solely on coarse-grained supervision from the spatial localization loss to learn prompts, with a lack of fine-grained object cues feedback (Fig. \ref{fig:difference_a})}, they inevitably generate vague prompts that struggle to clearly pinpoint specific object information in nighttime aerial scenarios, impairing the tracker's ability to distinguish reliable cues and localize the object on the ground \cite{c:chow2025prompt}. As a result, the trackers still perform poorly when encountering darkness.

Motivated by the coupling property of prompts with object attributes \cite{c:jia2022visual}, we propose a novel nighttime aerial tracker called \textbf{DPTrack}, which adopts the \textbf{D}irectional kernel-guided \textbf{P}rompt learning for robust \textbf{Track}ing, as shown in Fig. \ref{fig:difference_b}. DPTrack encodes the given, yet often-overlooked, object template's specific features into the directional kernel (DK), which serves as the theoretically validated fine-grained guidance signals for prompt generation and efficiently improves the tracker's perceptual capability through an effective design. Specifically, we first mimic the hierarchical perception mechanism of the human visual system \cite{c:lou2025overlock,r:chen2025yolo}, designing the dual particle perception module (DPP) to capture local–global topological relationships in the target features and to strengthen its representation through cross-particle fusion of attribute features. Subsequently, we construct the direction-kernel adaptive encoder (DKE) to encode the topology-aware features into the directional kernel with fine-grained cues, whose theoretically validated directional selectivity serves as guidance for prompt generation. Finally, based on the attributes of the feature’s channel-category correspondence \cite{c:li2019siamrpn++}, we propose the kernel-guided prompt module (KGP), which propagates the kernel across the features of the search region, employs channel-wise affinities to indicate the positions of target features, and maps them into positional prompts derived from closed-form statistics through L2 normalization, a parameter-free process mitigating the uncertainty inherent in dynamic attention, guiding the tracker to accurately focus the object.

In summary, the contributions of this paper are as follows:
\begin{IEEEenumerate}
\item{We propose DPTrack, a novel prompt-based aerial tracker featuring the first guidance mechanism that leverages fine-grained cues derived from intrinsic attributes to generate high-quality prompts for accurate nighttime aerial tracking.} 
\item{We design an object-specific prompt optimization strategy based on the directional kernel, which exploits bionic perception and the kernel’s channel-affinity attributes to generate precise prompts that enable the perceived features to semantically point to the object.}
\item{Extensive experiments on five benchmarks show that DPTrack achieves superior performance (e.g., a 4.3\% improvement in average tracking precision on the UAVDark135 benchmark \cite{r:li2022all}), significantly outperforming existing state-of-the-art (SOTA) trackers.}
\end{IEEEenumerate}

\section{Related Work}
\subsection{Object Tracking in Aerial Image.}Existing aerial trackers can be categorized into two groups: early correlation filter-based trackers and  template similarity matching-based trackers.
\subsubsection{Correlation filter-based trackers}Correlation filter-based trackers learn a discriminative filter and use Fourier correlation to locate the target. DSST \cite{r:dsst2016} learns translation and single dimension scale filters for size variations, but its simplified scale modeling compromises robustness against appearance changes and fast motion. STRCF \cite{c:strcf2018} incorporates temporal regularization into spatially regularized filters and solves it with ADMM for real-time tracking, but its performance remains sensitive to parameter tuning and large appearance changes. ARCF \cite{c:arcf} develops aberrance-repressed filters that exploit background patches and response-map regularization to mitigate boundary effects and occlusion-induced noise in aerial scenario tracking. AutoTrack \cite{c:autotrack} adopts automatic spatio-temporal regularization from local and global response variations to adaptively adjust spatial constraints and filter updates. IBRI\cite{r:ibri} leverages interval-based response inconsistency for multi-frame cues and a disruptor-aware scheme to suppress occlusions and distractors. RACF \cite{r:racf} introduces residue-aware correlation filters that integrate spatial–temporal regularization for frame-to-frame consistency and object scale refinement for size adaptation, thereby improving the robustness and accuracy of aerial tracking.
\subsubsection{Template similarity matching-based trackers}Template matching-based trackers locate object by comparing similarity between template and search regions. HiFT \cite{c:hift} introduces a hierarchical feature transformer that fuses shallow spatial and deep semantic cues across layers. TCTrack \cite{c:tctrack} employs temporally adaptive convolution to calibrate weights using historical frames and a temporal transformer to refine similarity maps, while TCTrack++ \cite{r:tctrack++} enhances this with attention-based temporal adaptation and memory-efficient refinement. Aba-ViTrack \cite{c:abavit} adopts a one-stream ViT unifying feature learning and template-search coupling, with background-aware token halting to remove redundant tokens. AVTrack \cite{c:avtrack} selectively activates essential transformer blocks and learns view-invariant representations via mutual information maximization. ORTrack \cite{c:ortrack} applies spatial Cox process masking for occlusion-robust representation and adaptive knowledge distillation for compact deployment.
\subsection{Nighttime Tracking in Aerial Image.}  
\subsubsection{Nighttime Tracking with Low Light Enhancement}Researchers integrate low-light enhancement into tracking pipelines, enabling aerial trackers to recognize the target under nighttime scenarios. \cite{c:li2021adtrack} pioneered the use of a low-light enhancement module based on logarithmic transformation for brightness adjustment. However, direct brightness amplification risks noise magnification and artifacts. HighlightNet \cite{c:fu2022highlightnet} mitigated this by using local masks to selectively enhance pixels, suppressing external interference. Darklighter \cite{c:ye2021darklighter} applied the Retinex model \cite{r:ren2020lr3m} to decouple illumination-invariant features, but some critical low-intensity features may also be discarded. MambaTrack \cite{c:zhang2025mambatrack} adopts a dual enhancement strategy that fuses visual and linguistic information to effectively perceive object features under nighttime scenarios. While improving the adaptability of aerial trackers to nighttime scenarios, these methods often face a misalignment of optimization objectives between the enhancement modules and trackers, potentially overlooking tracking-relevant features \cite{c:ye2022unsupervised}, thereby reducing tracking stability. 
\subsubsection{Domain-adaptive framework for Nighttime Tracking}To improve tracker adaptability across domains, UDAT \cite{c:ye2022unsupervised} proposed an unsupervised domain adaptation framework with transformer bridging layers for feature alignment, enabling the transfer of tracking capability from daytime to nighttime. SAM-DA \cite{c:fu2024sam} harnesses the zero-shot generalization capability of SAM to align cross-domain features by constructing high quality training samples, significantly enhancing domain adaptation. DaDiff \cite{c:zuo2024dadiff} employs a diffusion-based progressive alignment paradigm with temporal scheduling to align nighttime and daytime features. PDST \cite{c:zhang2023progressive} applied progressive momentum updates for domain-style transfer, improving nighttime robustness by shifting source-domain styles. LVPTrack \cite{c:wu2025lvptrack} employs a teacher-student network for knowledge distillation and incorporates a voting mechanism to refine label alignment, mitigating the impact of noisy labels on tracking.
\subsubsection{Nighttime Tracking with Prompt Learning}NiDR \cite{r:lei2024nidr} employs channel-wise illumination sensitivity discrepancies to capture illumination-invariant representations and mitigate Retinex-induced artifacts \cite{c:cai2023robust}. Although NiDR does not explicitly adopt prompt learning, the differences in channel to illumination can be regarded as implicit prompts that guide the tracker toward salient features. DCPT \cite{c:zhu2024dcpt} adapts back-projection from super-resolution \cite{r:hu2023cycmunet+} to visual tracking, task-specific losses drive feature reconstruction to amplify local details as potent visual cues, enabling reliable object signature acquisition by daytime trackers. LTrack \cite{c:wang2024multi}  uses ideal illumination distributions as reference prompts, enforcing illumination-consistent responses in low-frequency semantic features through contrastive supervision, thereby adapting daytime trackers to nighttime conditions.

Prompt-based aerial trackers have made progress, but they only rely on coarse supervision and the absence of detailed guidance often yield suboptimal prompts, impairing localization precision at nighttime. To address this, we propose DPTrack, which introduces fine-grained signals to enhance tracking accuracy.
\begin{figure*}[t]
	\centering
	\includegraphics[width=1\textwidth]{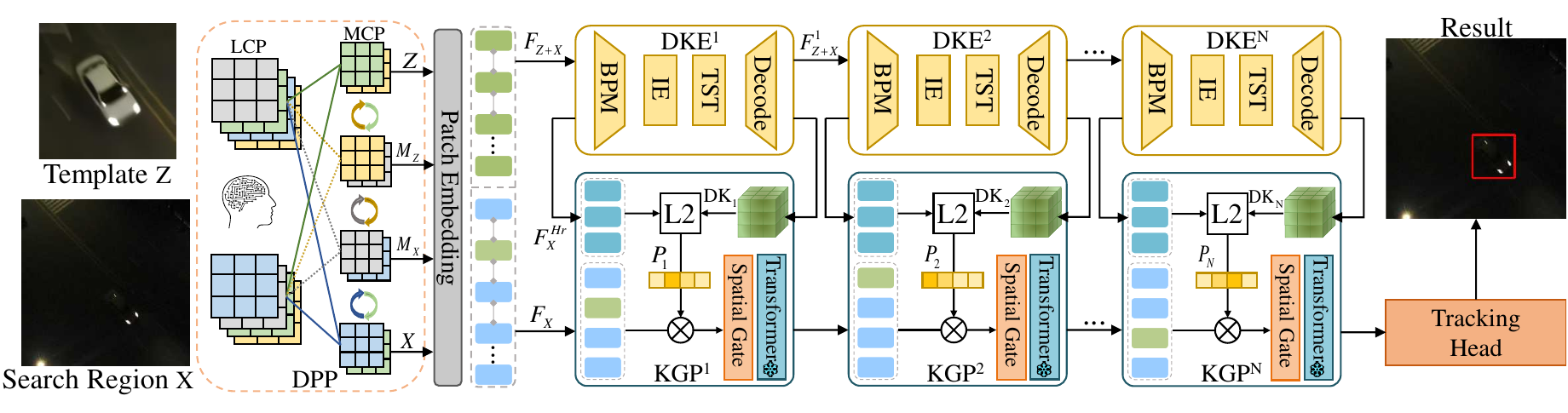}
	\caption{Overview of DPTrack. DPTrack utilizes DPP to establish global-local structural correlations, enhancing feature representation; DKE then transforms structured object information into the DK to guide accurate prompt generation. Finally, the KGP quantifies kernel correlations with search features to generate prompts that enable precise object localization.}
	\label{fig:DPTrack_arch}
\end{figure*}
\section{Proposed Method}
\begin{figure}[t]
	\centering
	\includegraphics[width=0.95\columnwidth]{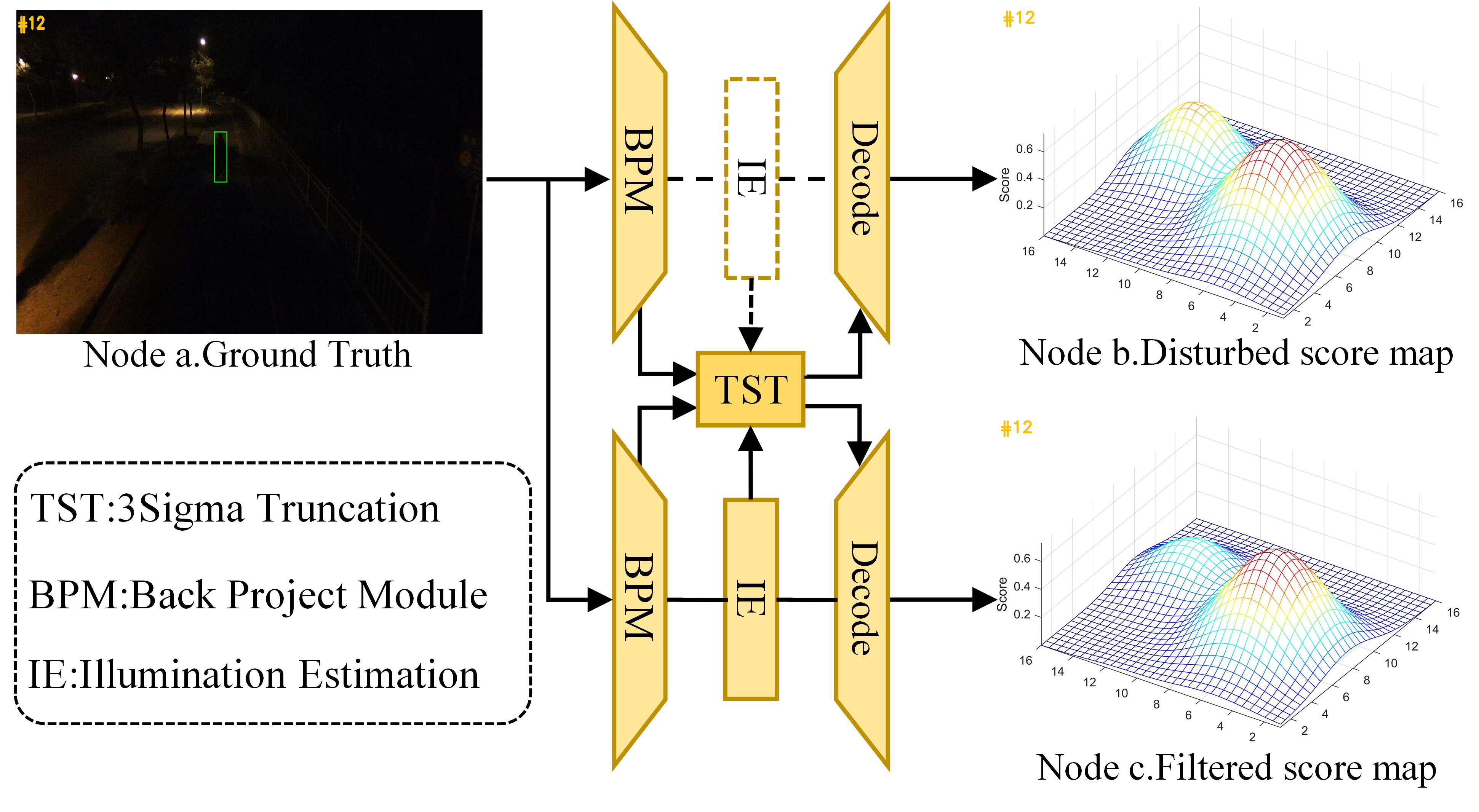} 
	\caption{The impact of non-uniform illumination on the score map: (a) The green ground-truth box indicates the object's true position, (b) presents the corresponding score map (artifact from localized glare) and (c) depicts the true score map after interference suppression by IE.}
	\label{fig:DKE}
\end{figure}

In this section, we provide a detailed introduction of DPTrack, as illustrated in Fig. \ref{fig:DPTrack_arch}. The DPTrack consists of three key components: (1) DPP (Fig. \ref{fig:DPP}) hierarchically extracts and correlates global-local structural features from template, significantly enhancing feature representation; (2) DKE (Fig. \ref{fig:DKE}) transforms structured features into the directional kernel that guides prompt generation; and (3) KGP (Fig. \ref{fig:KGP}) quantifies kernel correlations across the search region to generate positional prompts, enabling precise feature localization. To maintain feature symmetry between template $Z$ and search region $X$, both DPP and DKE are applied synchronously to $X$, following the protocol outlined by SiamRPN++ \cite{c:li2019siamrpn++}.
\subsection{Dual Particle Perception Module}
\begin{figure}[t]
	\centering
	\includegraphics[width=1\columnwidth]{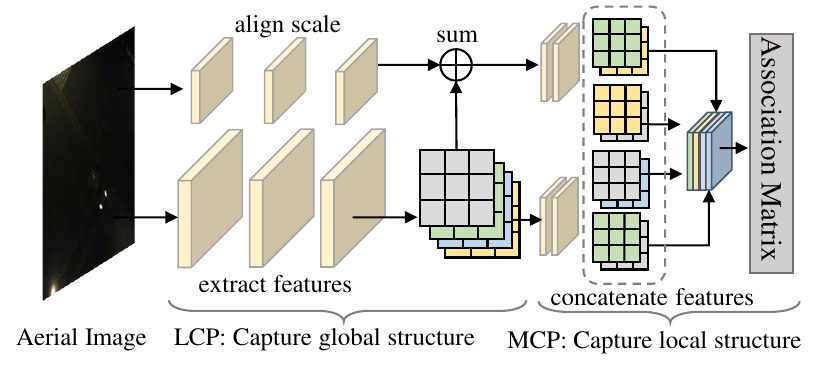} 
	\caption{The DPP module operates in two stages: LCP captures global structural information for scale-aligned concatenation with the original features, and MCP extracts local details to establish global-to-local structural associations. The fused features are then convolved to produce the association matrix. }
	\label{fig:DPP}
\end{figure}
The detailed structure of the DPP module is shown in Fig. \ref{fig:DPP}, where it serves as a core component responsible for progressive perception in the DPTrack framework. Existing transformer-based trackers patch weak features, which disrupts geometric correlations \cite{c:heidari2023hiformer}, thereby diminishing the effectiveness of features used for constructing a reliable directional kernel, whereas the human visual system leverages hierarchical perception to build local feature correlations and capture robust features. This motivates DPP, which emulates the human visual system to strengthen global-local structural correlations through topological attributes. Specifically, DPP adopts the ``Overview-first-Look-Closely-next" hierarchical perception mechanism \cite{c:lou2025overlock}, using grouped perceptrons with equivalent large-kernel convolutions\cite{c:li2024shift, r:wang2025bid} to progressively establish multiscale topological feature relationships. Taking the template \(Z \in \mathbb{R}^{ 3\times H_Z \times W_Z} \) as an example, this process can be characterized as:
\begin{equation}
M_Z = \mathrm{LCP}(Z) + \mathrm{MCP}(Z + \mathrm{LCP}(Z)),
\end{equation}
where \(M_Z\) represents the correlation matrix, with LCP and MCP representing the macro-perceptor and micro-perceptor respectively. The LCP emulates the overview functionality of human vision, employing a  stacked multi-layer Conv–ReLU block to process nighttime aerial images and effectively capture broader global topological features across spatial regions. The MCP further implements a more detailed inspection by jointly taking $Z$ and the extracted global topological features as input, which shares the same overall architecture as the LCP but instead employs different kernels, performing fine-grained cross-scale convolutions. This coarse-to-fine process establishes bidirectional correlations between global topology and local structures through an association matrix \(M_Z\) \cite{c:cai2023robust}, where \(M_Z\) and the input features are jointly embedded into a shared latent space in order to enrich the representation, which can be formally characterized as the following formulation:
\begin{equation}
	F_{Z+X}=\mathrm{PatchEmbed}([M_Z+Z;M_X+X]),
\end{equation}
where \(F_{Z+X} \in \mathbb{R}^{2C \times H \times W}\) denotes the structured features, $C$ represents the number of channels, $H$ and $W$ specify the spatial dimensions, and [;] refers to concatenation.
\subsection{Directional Kernel Adaptive Encoder}
To encode the \(F_{Z+X}\) into the \(DK\) and align it with the features of each layer, we design an encoder called DKE, as shown in Fig. \ref{fig:DKE}. Formally, DKE consists of a back projection module (BPM) \cite{c:zhu2024dcpt} and a third-order standard deviation truncation filter (TST).
The pathway from node a to b details the original architecture of DKE, the dashed lines in the flow indicate the absence of this processing step.
However, low-light scenarios are often accompanied by uneven illumination, where localized glare causes certain regions to exhibit brightness far exceeding the average. Such interference disrupts the effective modulation of feature strength by the BPM, causing the key feature locations decoded by the tracking head to become insufficiently emphasized and thereby generating false cues, such as Node b, ultimately introducing bias during the constrained prompt generation process, the prompts generated by the directional kernel that retains glare information cause noticeable interference in the tracker's object localization.

To address the interference effects, we design the Illumination Estimation module (IE), and incorporate it into the BPM to adaptively suppress the interference (Node c). Its processing procedure can be described as follows:
\begin{equation}
	F_{Z+X}^{Hr} = \alpha \mathrm{IE^1}(F_P^1) + \mathrm{FP^2}\left( \mathrm{IE^2}\left( F_D^1 - \beta F_{Z+X} \right) \right), \\
\end{equation}
\begin{equation}
	F_P^1 = \mathrm{FP^1}(F_{Z+X}), \quad
	F_D^1 = \mathrm{FD^1}\left( \mathrm{IE^1}(F_H^1) \right),
\end{equation}
where  \(\alpha,\beta \in \mathbb{R}^{1 \times 1}\) are learnable parameters, FP and FD denote the feature upsample function, feature downsample function of BPM. 
The process differs from baseline in that: the input \(F_{Z+X}\) first passes through the  FP to emphasize object cues \(F^1_{P} \in \mathbb{R}^{2C \times 2H \times 2W}\), then the IE suppresses interference by estimating the global brightness, and the FD restores feature details. The difference between them provides feedback on the error of the FP. The next IE modulates this error and uses the FP to compensate for it. Finally, a residual connection enhances object cues while preserving the original features.

Although DKE captures object cues, redundant background information is retained. To address this, we design a TST, which generates a truncation mask based on the three-sigma rule to suppress background interference. Since the goal of DKE is to encode object features, TST exclusively filters template features \(F_Z^{Hr} \in \mathbb{R}^{C \times H_Z \times W_Z}\). Global average pooling is applied to compute the channel-wise mean  $F^n_{mean}$ of $F_Z^{Hr}$, $n$ indexes the channel and $\mu$, $\sigma$ denote the global mean and standard deviation, respectively. The channel-wise truncation mask is defined as $\mathbb{I}$:
\begin{equation}
	\mathbb{I}(F^n_{mean}) =
	\begin{cases}
		1, & \text{if } \; |F_{\text{mean}}^n - \mu| \leq 3\sigma \\
		0, & \text{else}
	\end{cases}
\end{equation}
the \(DK \in \mathbb{R}^{C \times H_Z \times W_Z}\) is formed by Hadamard product of the $\mathbb{I}$ and $F^{Hr}_Z$, filtering background noise:
\begin{equation}
	DK=F^{Hr}_{Z} \odot \mathbb{I}(F^n_{mean}),
\end{equation}
\begin{algorithm}[t]
	\caption{Stepwise Derivation of $S_{DK}(y)$}
	\label{alg:grad_sdk}
	\begin{algorithmic}[1]
		\setlength{\itemsep}{0pt}
		\setlength{\parskip}{0pt}
		\renewcommand{\baselinestretch}{0.6}\selectfont
		
		\STATE Define: 
		$LSE(y) = \tau \log \textstyle \sum_{k} e^{-\rho_k(y)/\tau} 
		\;\rightarrow\; S_{DK}(y).$ 
		\STATE Differentiate w.r.t.~$y$:
		\[
		\nabla_y S_{DK}(y) = s'(LSE(y))\, \nabla_y LSE(y).
		\]
		\STATE Derive the gradient:
		\[
		\nabla_y LSE(y) =
		\textstyle \sum_{k}
		\frac{e^{-\rho_k(y)/\tau}}
		{\sum_{j} e^{-\rho_j(y)/\tau}}
		\nabla_y \rho_k(y).
		\]
		\STATE $\rho_k(y) \;\rightarrow\; \text{Mahalanobis distance}$:
		\[
		\nabla_y S_{DK}(y) =
		s'(LSE(y))
		\textstyle \sum_{k}
		\omega_k(y)
		\frac{L^\top L(y - \eta_k)}{\rho_k(y)}.
		\]
	\end{algorithmic}
\end{algorithm}to verify that the designed directional kernel exhibits directional pointing attributes, we derive as follows. For clarity, the variable \(F_X^{Hr} \in \mathbb{R}^{C \times H_X \times W_X}\) is denoted by $y$. The feature selection formulation $S_{DK}$ based on $DK$ is expressed as:
\begin{equation}
	S_{DK}(y)=s(min(\rho_{DK}(y))),
\end{equation}
where $\rho_{DK}$ denotes the distance metric and $s$ is a strictly monotonically decreasing function, making the feature selection inversely correlated with $\rho_{DK}$. The non-differentiable min operator is replaced with the log-sum-exp (LSE), whose differentiation process is detailed in Algorithm~\ref{alg:grad_sdk}. The Step-3 expression applies to any differentiable distance metric, and $\rho_{DK}$ is instantiated as the Mahalanobis distance with the mapping matrix $L$. As $s' < 0$, the result simplifies to:
\begin{equation}
	\label{eq:8}
	\nabla_y S_{DK}(y) \;\propto\; 
	- \sum\nolimits_{k} \omega_k(y)\,\frac{L^\top L\,(y-\eta_k)}{\rho_k(y)},
\end{equation}where $\eta_k$ is the subset of the \(DK\). From Eq. (\ref{eq:8}), each component points from $\eta_k$ towards $y$. Since the gradient ascent direction consistently drives $y$ towards the nearest prototype. In this way, the directional kernel is ensured to peak at the best-matching location and to offer local guidance toward the prototype, demonstrating its directional selectivity.

\begin{figure*}[t]
	\centering
	\includegraphics[width=1\textwidth]{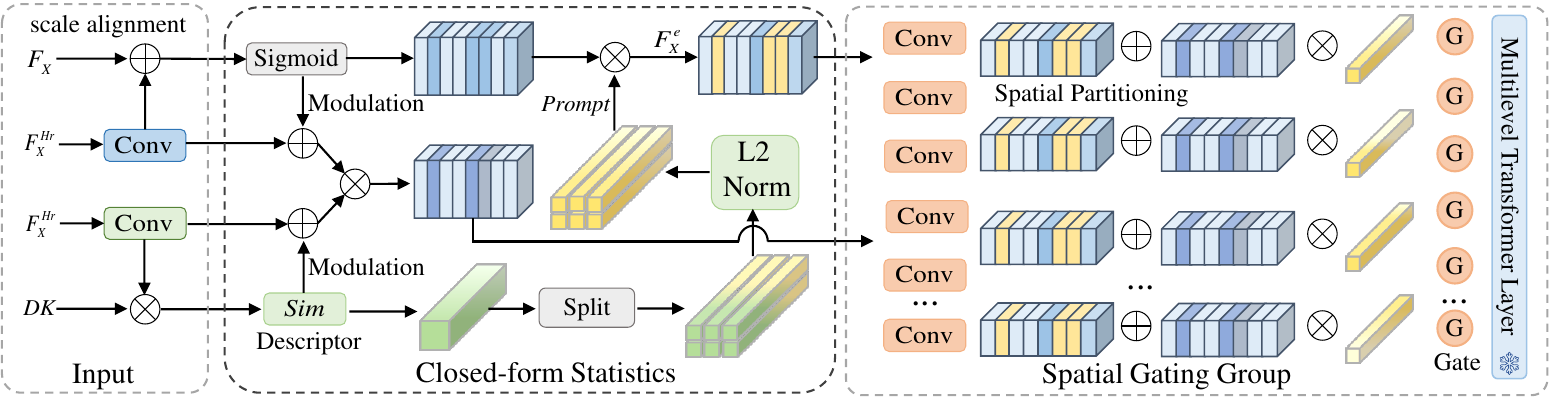} 
	\caption{Overview of the KGP pipeline. Input features are scale-aligned by convolution, and two separate convolutions are applied to adjust $F^{Hr}_X$ for differences in emphasized dimensions, producing a fused representation with $F_X$ that preserves the original feature information. The $DK$ estimates the confidence of each channel in $F^{Hr}_X$ for object indication, and the refined $Sim$, after L2 normalization, is employed as prompt to guide the tracker toward high-confidence channels. Finally, gating units partition the features along the spatial dimension to filter out noise.}
	\label{fig:KGP}
\end{figure*}
\subsection{Kernel-Guided Prompt Module}
The KGP computes channel-wise feature affinities based on the channel-category correspondence attributes \cite{c:li2019siamrpn++}. These affinities are L2-normalized and reconstructed into the prompt $P$, indicating the per-channel confidence of object presence, and guiding the tracker's attention toward high-confidence channels. In contrast to conventional attention, KGP generates prompts through a parameter-free closed-form statistical paradigm, mitigating uncertainty and the risk of overfocusing on irrelevant features. The KGP is illustrated in Fig. \ref{fig:KGP}, which performs dimensional expansion to align the features of \(DK\) with \(F_X^{Hr} \in \mathbb{R}^{C \times H_X \times W_X}\), and then applies cross-correlation to quantify the affinity between the template and the search region, as follows:
\begin{equation}
	Sim =
	 \sum_{h=0}^{H-1} \sum_{w=0}^{W-1}DK(c_1+\Delta c,h,w)\, F_X^{Hr}(c_2,h,w),
\end{equation}
where \(Sim \in \mathbb{R}^{C \times 1 \times 1}\) denotes the channel-wise descriptor, $\Delta_c=c_2-c_1$. The channel descriptor $Sim_c$ quantifies the confidence of object features in the $c$-th channel of $F^{Hr}_X$. This provides guidance weights for the tracker. Since the energy distribution across channels reflects their categorical differences, DPTrack adopts L2 normalization instead of sigmoid to preserve the consistency of inter-channel differences: 
\begin{equation}
	F^e_X = (\frac{Sim}{\sqrt{Sim_1^2 + \cdots + Sim_c^2 + \varepsilon}}+1)\times (F^{Hr}_X+F_X)
\end{equation}

Normalizing $Sim$ yields the $P$, which denotes the object positioning prompt, and adaptively highlight the region relevant features \(F^e_X \in \mathbb{R}^{C \times H_X \times W_X}\), guiding the tracker's attention to discriminative features. 
Since features may contain intra-class distractors, channel-wise feature modulation can mistakenly activate similar interference. Therefore, we adopt the spatial gate \cite{c:rama2024gated} to suppress such interference. Formally, the mechanism consists of cascaded spatial gating units, which partition the features along the channel dimension and regulate the spatial distribution of intra-channel features through learnable gating weights:
\begin{equation}
	F_{out} = \sum_{\xi\in\Omega}(g_n(\xi) \times F_X^e + (1 - g_n(\xi)) \times F_X^{Hr}), 
\end{equation}
where $g_n(\xi)$ denotes the $n$-th learnable gating unit of the direct mapping,  $\Omega$ represents the feature space, and \(\xi \in \mathbb{R}^{1 \times 1}\) indicates the learning coefficient. This mechanism serves as a spatial feature selector, performing adaptive allocation of spatial weights in collaboration with channel-wise prompts to emphasize fine-grained object cues. In this way, the spatial gating mechanism dynamically suppresses interference from redundant or confusing features thereby improving discriminability and robustness compared with conventional static gating formulations.
\subsection{Training objective}
Since nighttime aerial trackers share identical training objectives with general-purpose trackers, we employ the standard combination of L1 loss and GIoU loss to optimize localization:
\begin{equation}
	\mathcal{L}_{locate} = \lambda_1 \mathcal{L}_1(B_{pr}, B_{gt}) + \lambda_G \mathcal{L}_{GIOU}(B_{pr}, B_{gt}),
\end{equation}
where $B_{pr}$ denotes the predicted coordinates, $B_{gt}$ is the ground truth, $\lambda_1$ and $\lambda_G$ are balancing weights. The joint optimization of coarse-grained loss constraints together with fine-grained guidance signals establishes an effective coarse-to-fine refinement mechanism.
\section{Experiments and Analysis}

\subsection{Experimental Setup}
\subsubsection{Datasets}In this section, five representative datasets were utilized: UAVDark135 \cite{r:li2022all}, NAT2021 \cite{c:ye2022unsupervised}, NAT2021-L \cite{c:ye2022unsupervised}, NAT2024 \cite{c:fu2024prompt}, and DarkTrack2021 \cite{r:ye2022tracker}. These datasets constitute the comprehensive benchmark for evaluating both the effectiveness and the generalization of our DPTrack under diverse nighttime aerial tracking scenarios. A detailed description of each dataset is provided below.
\begin{IEEEenumerate}
\item{NAT2021 \& NAT2021-L. NAT2021 \cite{c:ye2022unsupervised} is a challenging nighttime dataset containing 180 sequences and over 14k frames. Its subset, NAT2021-L, includes 23 extended sequences each containing more than 1,400 frames and is specifically designed for long-term evaluation.}

\item{UAVDark135. UAVDark135 \cite{r:li2022all} is a comprehensive dataset with 135 nighttime sequences and over 10K frames, featuring diverse scenes and rich object categories. Its meticulously verified and iteratively refined annotations provide a reliable benchmark for nighttime aerial tracking.}

\item{NAT2024-1. NAT2024-1 \cite{c:fu2024prompt} provides 100 meticulously annotated sequences covering diverse illumination conditions (night; dusk; dawn) and motion patterns. It is specifically designed to evaluate the robustness of trackers in low-light environments.}

\item{DarkTrack2021. DarkTrack2021 \cite{r:ye2022tracker} is a challenging benchmark with 110 nighttime sequences and over 11k frames. Captured at 30 FPS across diverse urban nighttime scenarios, it enables comprehensive performance evaluation under complex illumination conditions.}
\end{IEEEenumerate}

\subsubsection{Implementation Details}DPTrack consists of five modules: DPP, backbone, DKE, KGP, and head. The DPP comprises two identical convolutional blocks (LCP and MCP) connected by skip links, each containing three Conv–ReLU layers with kernel sizes of $5 \times 5$ and $3 \times 3$, respectively. The backbone adopts a ViT-256 model pre-trained on large-scale tracking datasets, with parameters frozen during training. The DKE integrates a back-projection structure with an IE module. Each projection stage in the back-projection structure incorporates three $3 \times 3$ convolutional layers with activation function, while the IE module applies two convolutional layers to estimate and normalize local illumination variations. The KGP is a parameter-free paradigm that converts affinity into positional prompts. Finally, the head adopts a corner-based design with two convolutional branches to regress the top-left and bottom-right coordinates of the object.

During training, we adopt a two-stage strategy of pre-training and fine-tuning. Template and search region images are cropped to $128 \times 128$ and $256 \times 256$, respectively. In pre-training, the backbone is initialized with DCPT \cite{c:zhu2024dcpt} weights and trained for 200 epochs with a batch size of 64 on LaSOT \cite{c:fan2019lasot}, GOT-10K \cite{r:huang2019got}, VID \cite{r:russakovsky2015imagenet}, and COCO \cite{c:lin2014microsoft}, following the classical paradigm. Fine-tuning is performed exclusively on nighttime data from BDD100K \cite{c:yu2020bdd100k}, SHIFT \cite{c:sun2022shift}, ExDark \cite{r:loh2019getting}, and LaSOT\cite{c:fan2019lasot}, with a sampling ratio of $2:2:3:2$. At this stage, the backbone is frozen and the model is optimized for 80 epochs using AdamW with step decay after 48 epochs. To enhance generalization, we apply data augmentation including probabilistic grayscaling, random flipping, and jitter-based box perturbations, expanding the number of training samples per epoch to 60000, which increases spatial diversity and better simulate real-world variations. The implementation is based on Python 3.9 and PyTorch 1.13 on Ubuntu 20.04, with training conducted on dual NVIDIA RTX 3090 GPUs.

The inference configuration remains consistent with the training phase, and all experimental results are reported on a percentage scale, while all runs are executed on a single GPU to simulate offline deployment.

\subsubsection{Evaluation Metrics}We adopt widely used evaluation metrics to assess both the performance and the complexity of the proposed method, including the area under the success curve (AUC), precision (Prec.), normalized precision(Norm. Prec.), frames per second (FPS), floating-point operations (FLOPs), and the number of parameters (Params). The definitions of these performance metrics are provided as follows:

\begin{equation}
\mathrm{AUC} = \int_{0}^{1} \frac{1}{T} \sum_{t=1}^{T} 
\mathbf{1}\!\left( 
\frac{\left| B_{pr}^{t} \cap B_{gt}^{t} \right|}
{\left| B_{pr}^{t} \cup B_{gt}^{t} \right|}
> \zeta \right) d\zeta ,	
\end{equation}
where $B_{pr}^{t}$ and $B_{gt}^{t}$ denote the predicted bounding box and the ground-truth at frame $t$. $|\cdot|$ represents the area of a region, $\cap$ and $\cup$ denote intersection and union operations, respectively. 
$T$ is the total number of frames, $\zeta \in [0,1]$ is the overlap threshold, and $\mathbf{1}(\cdot)$ is the indicator function that outputs $1$ if the condition is satisfied and $0$ otherwise.
\begin{equation}
	\mathrm{Prec}(\phi) = 
	\frac{1}{T} \sum_{t=1}^{T} 
	\mathbf{1}\!\left( 
	\| C^t_{pr} - C^t_{gt} \|_2 \leq \phi
	\right),
\end{equation}
where $C^t_{pr}$ and $C^t_{gt}$ denote the center coordinates of the predicted and ground-truth bounding boxes, respectively. $\phi$ is the distance threshold. Norm. Prec. represents the scale-normalized precision metric. FPS, FLOPs, and Params are reported using PyTorch toolkits.

\subsection{Comparison With State-of-the-Arts}
\begin{table*}[t]
	\centering
	\renewcommand{\arraystretch}{1.1}
	\caption{Quantitative performance comparison of SOTA nighttime aerial trackers and DPTrack on the UAVDark135 dataset. The top three results are highlighted in bold and underlined. Upward arrows indicate that higher values correspond to better performance.}
	\resizebox{\textwidth}{!}
	{
		\begin{tabular}{l|l|c|c|c|c|c|c|c}
			\hhline{=========}
			Trackers & \rule{0pt}{1em}Venue & Prec. $\uparrow$ & $\Delta_{Prec.}$ & Norm. Prec. $\uparrow$ & $\Delta_{Norm.}$ & AUC $\uparrow$ & $\Delta_{AUC}$ & Speed (FPS) \\
			\hline
			Ocean\cite{c:ocean}       & ECCV'20  & 43.6 & -31 & 43.0 & -30.9 & 34.2 & -26.6 & 91.4 \\
			PRDIMP50-SCT\cite{c:prdimp50} & CVPR'20 & 66.7 & -7.9 & 66.5 & -7.4 & 52.8 & -8.0 & 31.25 \\
			SiamAPN\cite{c:siamapn}     & ICRA'21  & 42.2 & -32.4 & 40.8 & -33.1 & 30.6 & -30.2 & 143 \\
			HiFT-SCT\cite{c:hift}    & ICCV'21  & 53.8 & -20.8 & 53.8 & -20.1 & 41.0 & -19.8 & 43.7 \\
			UDAT-BAN\cite{c:ye2022unsupervised}    & CVPR'22  & 61.3 & -13.3 & 60.0 & -13.9 & 47.2 & -13.6 & 46 \\
			DeconNet\cite{r:deconnet}    & TGRS'22  & 48.3 & -26.3 & 47.7 & -26.2 & 38.7 & -22.1 & 131.7 \\
			UDAT-CAR\cite{c:ye2022unsupervised}    & CVPR'22  & 60.7 & -13.9 & 61.2 & -12.7 & 48.5 & -12.3 & 46.4 \\
			MAT\cite{c:mat}         & CVPR'23  & 57.2 & -17.4 & 57.6 & -16.3 & 47.1 & -13.7 & 56 \\
			HiT-Base\cite{c:hitbase}    & ICCV'23  & 48.9 & -25.7 & 48.7 & -25.2 & 41.1 & -19.7 & 156 \\
			Aba-ViTrack\cite{c:abavit} & ICCV'23  & 61.3 & -13.3 & 63.5 & -10.4 & 52.1 & -8.7 & 134 \\
			TCTrack++\cite{r:tctrack++}   & TPAMI'23 & 47.4 & -27.2 & 47.4 & -26.5 & 37.8 & -23 & 27.1 \\
			AVTrack-DeiT\cite{c:avtrack} & ICML'24 & 58.6 & -16.0 & 59.2 & -14.7 & 47.6 & -13.2 & 212 \\
			NiDR\cite{r:lei2024nidr}        & TGRS'24  & 64.2 & -10.4 & 62.9 & -11.0 & 51.1 & -9.7 & 71.6 \\
			DCPT\cite{c:zhu2024dcpt}        & ICRA'24  & \underline{70.3} & -3.3 & \underline{70.1} & -3.8 & \underline{57.7} & -3.1 & 60 \\
			DARTer\cite{c:darter}      & ICMR'25  & \underline{71.6} & -3.0 & \underline{72.1} & -1.8 & \underline{58.2} & -2.6 & 71.6 \\
			MCITrack\cite{c:kang2025exploring}    & AAAI'25  & 67.6 & -7.0 & 61.6 & -12.3 & 56.0 & -4.8 & 35 \\
			ORTrack\cite{c:ortrack}    & CVPR'25  & 59.6 & -15.0 & 60.4 & -13.5 & 48.6 & -12.2 & 119 \\
			SGLATrack-DeiT\cite{c:sglatrack}    & CVPR'25  & 63.8 & -10.8 & 64.2 & -9.7 & 51.9 & -8.9 & 135 \\
			\textbf{DPTrack} & Ours & \textbf{74.6} & - & \textbf{73.9} & - & \textbf{60.8} & - & 49 \\
			\hhline{=========}
		\end{tabular}
		\label{tab:results_uavdark135}
	}
\end{table*}

\begin{figure*}[t]
	\centering
	\includegraphics[width=1\textwidth]{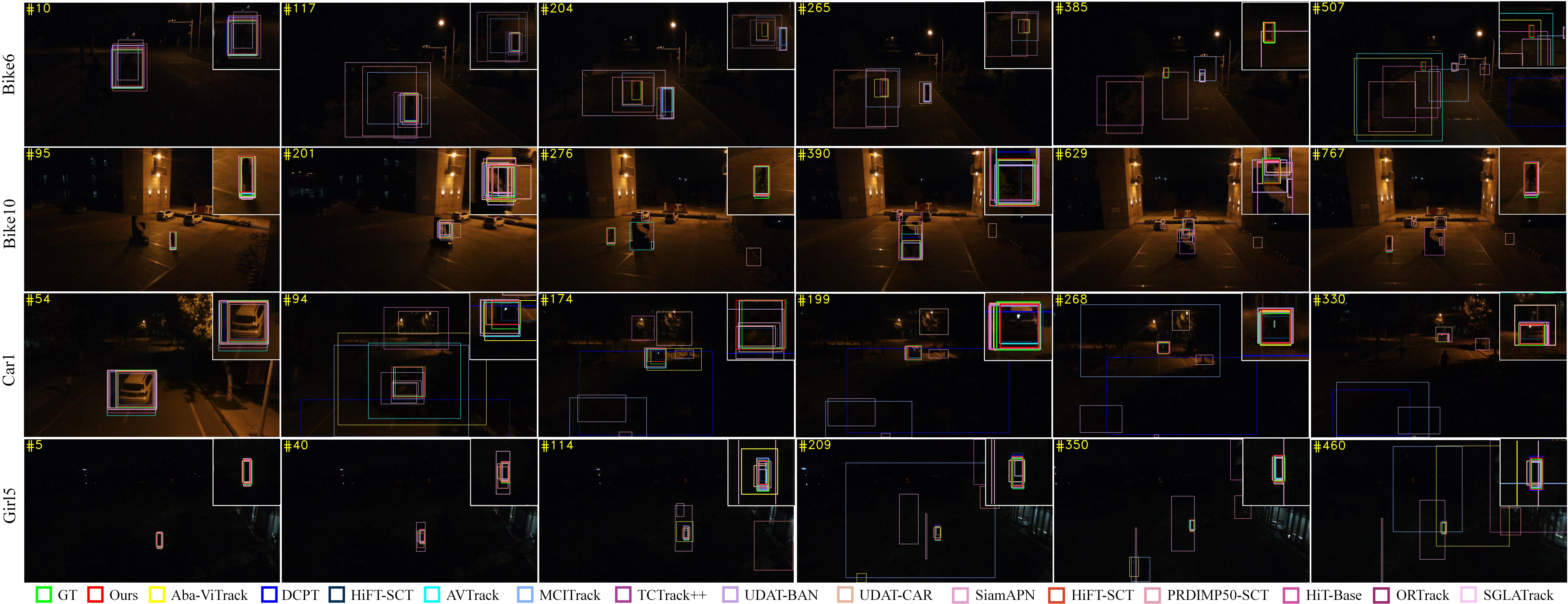}
	\caption{Qualitative evaluation of SOTA trackers and DPTrack on the UAVDark135 benchmark. Representative sequences are visualized, where the ground truth is shown in green and DPTrack is highlighted in red.}
	\label{fig:vis_uavdark135}
\end{figure*}

\begin{table*}[t]
	\centering
	\caption{Quantitative performance comparison of SOTA nighttime aerial trackers and DPTrack on the NAT2021 dataset. The top three results are highlighted in bold and underlined. Upward arrows indicate that higher values correspond to better performance.}
	\resizebox{\textwidth}{!}
	{
		
		\begin{tabular}{l|l|c|c|c||c|c|c|c|c}
			\hhline{==========}
			Trackers & \rule{0pt}{1em}Prec. $\uparrow$ & $\Delta_{Prec.}$ & AUC $\uparrow$ &
			$\Delta_{AUC}$ & Trackers & Prec. $\uparrow$ &
			$\Delta_{Prec.}$ & AUC $\uparrow$ & $\Delta_{AUC}$ \\
			\hline
			Ocean\cite{c:ocean}    & 58.9 & -11.8 & 38.9 & -14.8 & MAT\cite{c:mat} & 64.8 & -5.9 & 47.7  & -6.0 \\
			TCTrack\cite{c:tctrack}    & 60.8 & -9.9 & 40.8 & -12.9 & HiT-Base\cite{c:hitbase} & 49.3 & -21.4 & 36.4  & -17.3 \\
			HiFT-SCT\cite{c:hift} & 60.6  & -10.1 & 41.7 & -12.0 & AVTrack-DeiT\cite{c:avtrack} & 61.5 & -9.2 & 45.5 & -8.2 \\
			DeconNet\cite{r:deconnet}         & 63.7  & -7.0 & 43.9 & -9.8 & NiDR\cite{r:lei2024nidr} & 65.2 & -5.5 & 47.0 & -6.7 \\
			UDAT-CAR\cite{c:ye2022unsupervised} & \underline{68.2}  & -2.5 & 48.7 & -5.0 & DCPT\cite{c:zhu2024dcpt} & \underline{69.0} & -1.7 & \underline{52.6} & -1.1 \\
			Aba-ViTrack\cite{c:abavit}  & 60.4  & -10.3 & 46.9 & -6.8 & MCITrack\cite{c:kang2025exploring} & 66.4 & -4.3 & \underline{53.0} & -0.7 \\
			ORTrack\cite{c:ortrack}  & 65.1  & -5.6 & 48.0 & -5.7 & SGLATrack-DeiT\cite{c:sglatrack} & 64.8 & -5.9 & 48.2 & -5.5 \\
			TCTrack++\cite{r:tctrack++}    & 61.6 & -9.1 & 41.7 & -12.0 & \textbf{DPTrack} & \textbf{70.7} & - & \textbf{53.7}  & - \\
			\hhline{==========}
		\end{tabular}
		\label{tab:results_nat2021}
	}
\end{table*}

\begin{table*}[t]
	\centering
	\caption{Quantitative performance comparison of SOTA nighttime aerial trackers and DPTrack on the NAT2024-1 dataset. The top three results are highlighted in bold and underlined. Upward arrows indicate that higher values correspond to better performance.}
	\resizebox{\textwidth}{!}
	{
		\begin{tabular}{l|l|c|c|c||c|c|c|c|c}
			\hhline{==========}
			Trackers & \rule{0pt}{1em}Prec. $\uparrow$ & $\Delta_{Prec.}$ & AUC $\uparrow$ &
			$\Delta_{AUC}$ & Trackers & Prec. $\uparrow$ &
			$\Delta_{Prec.}$ & AUC $\uparrow$ & $\Delta_{AUC}$ \\
			\hline
			SGDViT\cite{c:sgdvit}    & 53.1 & -30.6 & 38.1 & -26.4 & MAT\cite{c:mat} & 80.5 & -3.2 & 61.9  & -2.6 \\
			TCTrack\cite{c:tctrack}    & 74.4 & -9.3 & 47.0 & -17.5 & HiT-Base\cite{c:hitbase} & 62.7 & -21.0 & 48.2  & -16.3 \\
			HiFT-SCT\cite{c:hift} & 60.6  & -23.1 & 41.4 & -23.1 & AVTrack-DeiT\cite{c:abavit} & 75.3 & -8.4 & 56.7 & -7.8 \\
			TDA-Track\cite{c:tdatrack}         & 75.5 & -8.2 & 51.4 & -13.1 & LiteTrack\cite{c:wei2024litetrack} & \underline{82.4} & -1.3 & \underline{62.7} & -1.8 \\
			UDAT-CAR\cite{c:ye2022unsupervised} & 69.8  & -13.9 & 50.6 & -13.9 & DCPT\cite{c:zhu2024dcpt} & 81.1 & -2.6 & 62.1 & -2.4 \\
			Aba-ViTrack\cite{c:abavit}  & 78.4  & -5.3 & 60.1 & -4.4 & MCITrack\cite{c:kang2025exploring} & \underline{81.4} & -2.3 & \underline{64.5} & 0.0 \\
			ORTrack\cite{c:ortrack}  & 81.5  & -2.2 & 61.3 & -3.2 & SGLATrack-DeiT\cite{c:sglatrack} & 73.6 & -10.1 & 56.2 & -8.3 \\
			TCTrack++\cite{r:tctrack++}    & 70.5 & 13.2 & 46.6 & -17.9 & \textbf{DPTrack} & \textbf{83.7} & - & \textbf{64.5}  & - \\
			\hhline{==========}
		\end{tabular}
		\label{tab:results_nat2024}
	}
\end{table*}

\begin{table*}[t]
	\centering
	\caption{Quantitative performance comparison of SOTA nighttime aerial trackers and DPTrack on the DarkTrack2021 dataset. The top three results are highlighted in bold and underlined. Upward arrows indicate that higher values correspond to better performance.}
	\resizebox{\textwidth}{!}
	{
		\begin{tabular}{l|l|c|c|c||c|c|c|c|c}
			\hhline{==========}
			Trackers & \rule{0pt}{1em}Prec. $\uparrow$ & $\Delta_{Prec.}$ & AUC $\uparrow$ &
			$\Delta_{AUC}$ & Trackers & Prec. $\uparrow$ &
			$\Delta_{Prec.}$ & AUC $\uparrow$ & $\Delta_{AUC}$ \\
			\hline
			Ocean\cite{c:ocean}    & 53.9 & -14.2 & 40.9 & -15.9 & SiamRPN++\cite{c:li2019siamrpn++} & 50.9 & -17.2 & 38.6  & -16.6 \\
			SiamAPN\cite{c:siamapn}    & 42.4 & -25.7 & 31.4 & -23.8 & TCTrack\cite{c:tctrack} & 54.8 & -13.3 & 40.8 & -14.4 \\
			HiFT-SCT\cite{c:hift} & 53.5  & -14.6 & 42.6 & -12.6 & LiteTrack\cite{c:wei2024litetrack} & \underline{67.6} & -0.5 & 54.3 & -0.9 \\
			DeconNet\cite{r:deconnet}         & 56.0  & -12.1 & 42.7 & -12.5 & NiDR\cite{r:lei2024nidr} & 61.7 & -6.4 & 48.0 & -7.2 \\
			UDAT-CAR\cite{c:ye2022unsupervised} & 59.9  & -8.2 & 47.0 & -8.2 & DCPT\cite{c:zhu2024dcpt} & 66.7 & -1.5 & \underline{54.0} & -1.2 \\
			TDA-Track\cite{c:tdatrack}  & 53.3 & -14.8 & 39.3 & -15.9 & MCITrack\cite{c:kang2025exploring} & \underline{66.9} & -1.2 & \underline{54.7} & -0.5 \\
			ORTrack\cite{c:ortrack}  & 60.5  & -7.6 & 48.6 & -5.1 & SGLATrack-DeiT\cite{c:sglatrack} & 58.8 & -9.3 & 48.0 & -7.2 \\
			TCTrack++\cite{r:tctrack++}    & 55.5 & -12.6 & 42.2 & -13.0 & 
			\textbf{DPTrack} & \textbf{68.1} & - & \textbf{55.2}  & - \\
			\hhline{==========}
		\end{tabular}
		\label{tab:results_darktrack2021}
	}
\end{table*}

\begin{table*}[h]
	\centering
	\caption{Quantitative performance comparison of SOTA nighttime aerial trackers and DPTrack on the NAT2021-L dataset. The top three results are highlighted in bold and underlined in table. Upward arrows indicate that higher values correspond to better performance.}
	\resizebox{\textwidth}{!}{
		\begin{tabular}{l|l|c|c|c||c|c|c|c|c}
			\hhline{==========}
			Trackers & \rule{0pt}{1em}Prec. $\uparrow$ & $\Delta_{Prec.}$ & AUC $\uparrow$ & $\Delta_{AUC}$ &
			Trackers & Prec. $\uparrow$ & $\Delta_{Prec.}$ & AUC $\uparrow$ & $\Delta_{AUC}$ \\
			\hline
			DCPT            & \underline{59.9} & -3.4  & \underline{47.4} & -2.2  & UDAT-BAN        & 49.0 & -14.3 & 35.4 & -14.2 \\
			LiteTrack       & \underline{58.2} & -5.1  & \underline{45.5} & -4.1  & TCTrack++        & 46.8 & -16.5 & 32.8 & -16.8 \\
			DIMP50-SCT     & 57.7 & -5.6  & 41.4 & -8.2  & SiamAPN++-SCT   & 46.0 & -17.3 & 32.2 & -17.4 \\
			SGLATrack       & 55.0 & -8.3  & 43.8 & -5.8  & SiamRPN-SCT     & 44.7 & -18.6 & 30.5 & -19.1 \\
			ORTrack         & 51.6 & -11.7 & 40.6 & -9.0  & HIFT-SCT        & 43.9 & -19.4 & 31.0 & -18.6 \\
			UDAT-CAR       & 49.7 & -13.6 & 35.8 & -13.8 & SiamAPN          & 38.4 & -24.9 & 24.2 & -25.4 \\
			TCTrack         & 48.0 & -15.3 & 30.7 & -18.9 & \textbf{DPTrack} & \textbf{63.3} & -- & \textbf{49.6} & -- \\
			\hhline{==========}
	\end{tabular}}
	\label{tab:results_nat2021l}
\end{table*}

\begin{figure*}[t]
	\centering
	\includegraphics[width=1\textwidth]{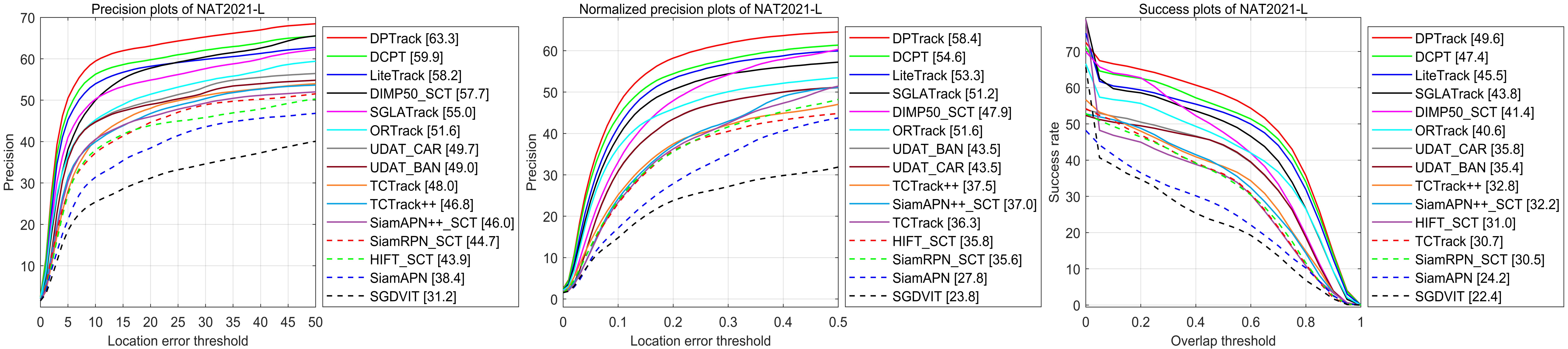}
	\caption{Comprehensive and intuitive evaluations of DPTrack and SOTA trackers on the NAT2021-L \cite{c:ye2022unsupervised} benchmark show that DPTrack consistently achieves the best performance across decision thresholds, demonstrating its superior ability to capture fine-grained target cues.}
	\label{fig:result_nat2021l}
\end{figure*}

\begin{figure*}[t]
	\centering
	\includegraphics[width=1\textwidth]{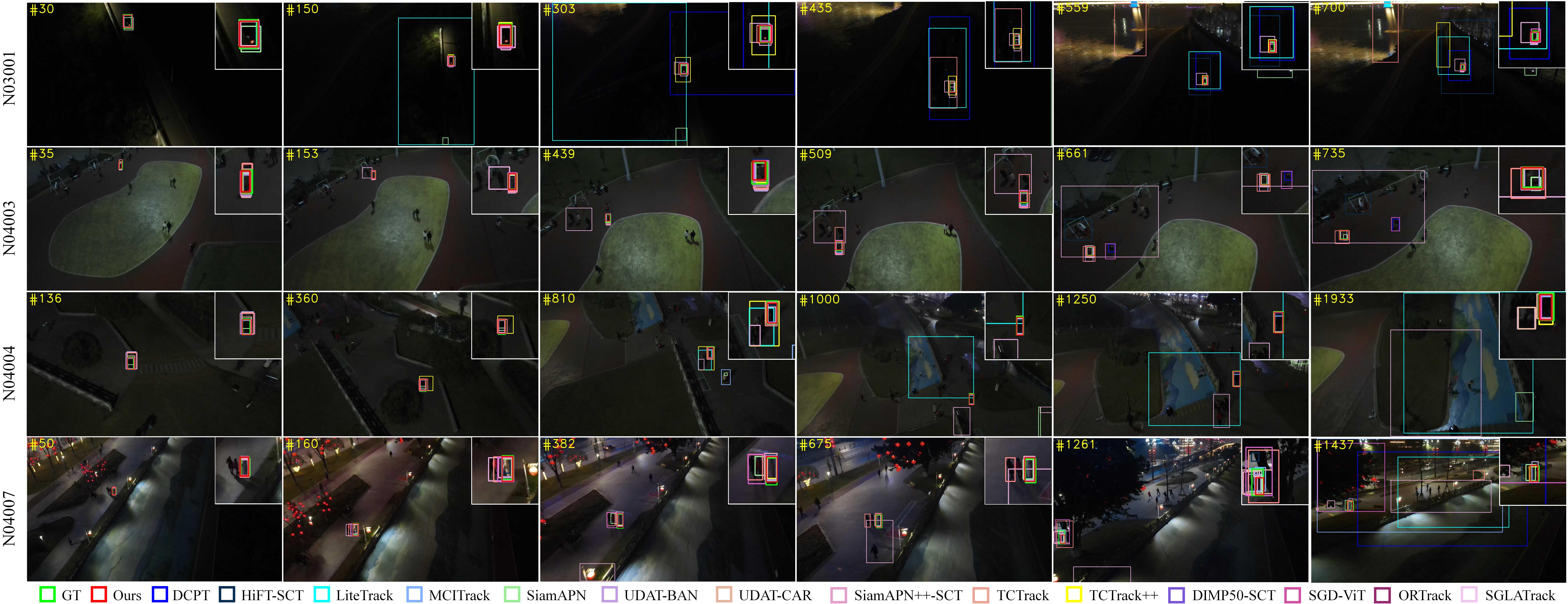}
	\caption{Qualitative evaluation of SOTA trackers and DPTrack on the NAT2021-L benchmark. Representative sequences are visualized, where the ground truth is shown in green and DPTrack is highlighted in red.}
	\label{fig:vis_nat2021l}
\end{figure*}

\subsubsection{Quantitative Evaluation}We quantitatively evaluate DPTrack on five benchmark datasets, summarized as follows:

\textbf{UAVDark135.} As shown in Table \ref{tab:results_uavdark135}, DPTrack achieves real-time performance while consistently outperforming 19 existing trackers across the three core metrics. Specifically, it surpasses DCPT \cite{c:zhu2024dcpt} by \textbf{4.3\%} in precision and \textbf{3.1\%} in AUC, and outperforms DARTer \cite{c:darter} by \textbf{3.0\%} in precision, \textbf{1.8\%} in normalized precision, and \textbf{2.6\%} in AUC, respectively. These results demonstrate the effectiveness of the proposed fine-grained guidance signals in alleviating environmental interference and highlighting target characteristics.

\textbf{NAT2021.} As shown in Table \ref{tab:results_nat2021}, the results on NAT2021 clearly demonstrate DPTrack's superior performance under various illumination interferences in urban environments, achieving a top level precision of \textbf{70.7\%} and AUC of \textbf{53.7\%}. It surpasses MCITrack \cite{c:kang2025exploring} by \textbf{4.3\%} in precision, NiDR \cite{r:lei2024nidr} by \textbf{5.5\%} in precision and \textbf{6.7\%} in AUC, confirming DPTrack's exceptional anti-interference capability and its consistent focus on the object across diverse complex scenarios.

\textbf{NAT2024-1.} DPTrack demonstrates remarkable adaptability to diverse scenarios in Table \ref{tab:results_nat2024}, consistently achieving the best performance over a wide range of SOTA trackers. Specifically, it exceeds AVTrack \cite{c:avtrack} by \textbf{8.4\%} in precision and \textbf{7.8\%} in AUC, highlighting its robustness against challenging tracking scenarios. Moreover, it also surpasses the scene-adaptive Aba-ViTrack \cite{c:abavit} by \textbf{5.3\%} in precision and \textbf{4.4\%} in AUC, further validating the effectiveness of DPTrack in handling complex motion patterns.

\textbf{DarkTrack2021.} On the DarkTrack2021 benchmark, we conduct a comprehensive comparison of DPTrack against 13 SOTA trackers, where DPTrack still achieves clearly superior performance. As shown in Table \ref{tab:results_darktrack2021}, DPTrack attains \textbf{68.1\%} precision and \textbf{55.2\%} in AUC, outperforming MCITrack \cite{c:kang2025exploring} at \textbf{66.9\%} in precision and \textbf{54.7\%} in AUC, TCTrack \cite{c:tctrack} at \textbf{54.8\%} in precision and \textbf{40.8\%} in AUC. Such consistent improvements highlight DPTrack’s remarkable capability to effectively mitigate the adverse impact of uneven illumination in nighttime visual tracking scenarios.

\textbf{NAT2021-L.} The evaluation Table \ref{tab:results_nat2021l} substantiate the overall superiority of DPTrack. To provide a more intuitive demonstration of DPTrack’s comprehensive advantages under different thresholds, we further assess its performance on long-term benchmark by curves. Long-term nighttime tracking is particularly challenging due to scale accumulation errors, which validates the accuracy of DPTrack in estimating target scales. On the NAT2021-L benchmark \cite{c:ye2022unsupervised}, DPTrack maintains the highest score across different thresholds and achieves the best AUC of \textbf{49.6\%}, outperforming the second- and third-ranked trackers by \textbf{2.2\%} and \textbf{4.1\%}, as shown in Fig. \ref{fig:result_nat2021l}.

\subsubsection{Qualitative Evaluation}As shown in Fig.~\ref{fig:vis_uavdark135} and Fig.~\ref{fig:vis_nat2021l}, we visualize representative results on UAVDark135 and NAT2021-L to qualitatively assess nighttime aerial tracking. DPTrack maintains stable localization under extremely low illumination, where existing trackers often confuse the background with the object or lose it entirely. In challenging scenes with background distractors (e.g., Bike10, N04003) or large scale variations (e.g., Car1, N04004), DPTrack produces compact and well-aligned bounding boxes, effectively adapting to object scale changes. Under extremely dark conditions (e.g., Girl5, N03001, N04007), it suppresses glare and accurately localizes the target, while others drift or mis-detect glare regions. These results demonstrate that DPTrack effectively distinguishes objects from background interference and achieves superior robustness in nighttime aerial tracking.
\subsection{Ablation Study}

\begin{table*}[t]
	\centering
	\renewcommand{\arraystretch}{1.1}
	\caption{Ablation study of DPTrack’s components and KGP’s normalization strategies. $\Delta$ denotes the performance gain.}
	\resizebox{\textwidth}{!}{
		\begin{tabular}{l|c|c|c|c||l|c|c|c|c}
			\hhline{==========}
			\multicolumn{5}{c||}{(a) DPTrack Components} & \multicolumn{5}{c}{(b) KGP Normalization Strategies} \\
			\hline
			Settings & Prec. $\uparrow$ & $\Delta_{Prec.}$ & AUC $\uparrow$ & $\Delta_{AUC}$ &
			Settings & Prec. $\uparrow$& $\Delta_{Prec.}$ & AUC $\uparrow$& $\Delta_{AUC}$ \\
			\hline
			Baseline                  & 59.9 & --   & 47.4 & --   & Softmax Normalization & 60.5 & -2.8 & 47.2 & -2.4 \\
			Baseline + DPP            & 60.7 & +0.8 & 47.9 & +0.5 & Sigmoid Normalization & 61.3 & -2.0 & 47.7 & -1.9 \\
			Baseline + DPP + DKE      & 62.4 & +2.5 & 48.8 & +1.4 & Min--Max Normalization & 60.2 & -3.1 & 47.4 & -2.2 \\
			Baseline + DKE + KGP      & 63.1 & +3.2 & 49.1 & +1.7 & L1 Normalization      & 61.0 & -2.3 & 48.6 & -1.0 \\
			\textbf{DPTrack}          & \textbf{63.3} & \textbf{+3.4} & \textbf{49.6} & \textbf{+2.2} &
			\textbf{L2 Normalization} & \textbf{63.3} & -- & \textbf{49.6} & -- \\
			\hhline{==========}
	\end{tabular}}
	\label{tab:ablation_combined}
\end{table*}

\subsubsection{Ablation Study of Components} We study the collaboration among the components. As DPP and KGP lack a direct bridge, their combination is excluded from the experiments:

\textbf{Baseline+DPP.} DPP employs equivalent large-kernel convolutions to capture object structural correlations. After integrating DPP into the baseline, the resulting model achieves \textbf{60.7\%} precision and \textbf{47.9\%} AUC (Table \ref{tab:ablation_combined}, left), with only a \textbf{1M} increase in parameters, these results validate DPP’s effectiveness in constructing global–local correlations and underscore its critical role in strengthening feature representation under challenging nighttime conditions.

\textbf{Baseline+DPP+DKE.} Integrating both DPP and DKE into the baseline, where fine-grained signals are directly fused with object features (Table \ref{tab:ablation_combined}, left), yielding \textbf{2.5\%} in precision and \textbf{1.4\%} improvement in AUC. These results confirm that the directional kernel effectively embeds object fine-grained cues into guidance signals. However, without the precise prompts from KGP, the tracker underperforms compared to DPTrack.

\textbf{Baseline+DKE+KGP.} DKE and KGP are crucial for DPTrack, as they enable fine-grained guided prompt generation. As shown in Table \ref{tab:ablation_combined}, integrating DKE and KGP into the baseline yields notable gains of \textbf{3.2\%} in precision and \textbf{1.7\%} in AUC. These results demonstrate their synergistic effect in generating accurate prompts, contributing to robust and stable tracking under nighttime conditions.

\subsubsection{Ablation Study of Normalization Strategies in KGP}
 We ablate normalization strategies for prompt generation in KGP (Table \ref{tab:ablation_combined}, right). Normalization strongly affects prompt discriminability and thus tracking accuracy. Specifically, Softmax emphasize a single dominant channel, while neglecting the auxiliary contributions of other channels, leading to performance degradation with only \textbf{60.5\%} in precision and \textbf{47.2\%} in AUC. Similarly, Sigmoid partially alleviates this issue by smoothing the weight distribution of channels, but it still distorts the spatial relationships and results in sub-optimal performance (\textbf{61.3\%} precision, \textbf{47.7\%} AUC). Min–Max normalization suffers from scale compression, suppressing inter-channel contrast and producing the lowest precision (\textbf{60.2\%}) among all settings. L1 normalization provides a more balanced prompts, yet the sparsity induced by projecting the weights onto a unit cube weakens the contributions of secondary features, causing the performance to remain inferior to L2 normalization (\textbf{61.0\%} precision, \textbf{48.6\%} AUC). L2 normalization preserves the  structural balance of channels, yielding the most discriminative prompts and achieving the best performance with \textbf{63.3\%} precision and \textbf{49.6\%} AUC. These results demonstrate the effectiveness of L2 normalization.
\subsubsection{Ablation Study of Loss Hyperparameters} We investigate the impact of training hyperparameters on tracker performance, focusing on the balance between loss weights ($\lambda_1$ and $\lambda_G$) and the number of IE ($N_{IE}$), as summarized in Table \ref{tab:para_ablation}. When $N_{IE}$ increases from 0 to 2 under a fixed configuration of $\lambda_1 = 2.0$ and $\lambda_G= 5.0$, both precision and AUC improve consistently, reaching the best performance at \textbf{63.3\%} precision and \textbf{49.6\%} AUC. This demonstrates that a moderate number of IE layers effectively estimates global brightness and suppresses glare interference on features, thereby contributing to robust target estimation. However, further increasing IE layers yields only marginal changes in precision (\textbf{63.0\%}) and a slight drop in AUC (\textbf{48.8\%}), suggesting redundancy and over-suppression. Reducing both $\lambda_1$ and $\lambda_G$ causes substantial degradation, with precision falling to \textbf{58.2–58.9\%} and AUC to \textbf{46.1–46.7\%}, indicating insufficient supervision. Conversely, emphasizing $\lambda_1$ excessively brings limited precision gains (\textbf{61.0–61.5\%}) but fails to reach the balanced configuration in AUC (\textbf{48.1–48.3\%}). Overall, the best trade-off is achieved when L1 and GIoU losses are jointly emphasized with moderate weighting and an appropriate number of IE layers, ensuring stable gains across both precision and robustness.
\subsubsection{Ablation Study of Dataset Ratio}
We further analyze the effect of dataset ratio on fine-tuning performance, as summarized in Table \ref{tab:para_ablation}, the dataset order is BDD100K, SHIFT, ExDark and LaSOT. When the four datasets are equally weighted ($1:1:1:1$), the tracker achieves \textbf{60.9\%} precision and \textbf{46.9\%} AUC, serving as a balanced baseline. Second, increasing the proportion of SHIFT data leads to minor improvements in AUC (up to \textbf{49.1\%}) but a noticeable decline in precision, indicating that the domain gap between synthetic SHIFT and real nighttime data limits the overall benefit. Overemphasizing BDD100K and SHIFT further degrades performance (\textbf{60.3\%} precision and \textbf{47.1\%} AUC), suggesting that excessive synthetic or day-oriented data disrupts the model’s adaptation to real data. Finally, the mixed ratio of $2:2:3:2$ achieves the best overall performance with \textbf{63.3\%} precision and \textbf{49.6\%} AUC, demonstrating that a carefully balanced dataset composition is crucial for enhancing both accuracy and robustness nighttime aerial tracking.
\begin{table}[h]
	\centering
	\caption{Ablation study of training hyperparameters. $\Delta$ denotes the performance gain.}
	\small
	\renewcommand{\arraystretch}{1.1} 
	\begin{tabularx}{\linewidth}
		{@{} r !{\vrule width 0.4pt} l *{4}{>{\centering\arraybackslash}X} @{}}
		\toprule
		\# & \textbf{$\lambda_1$ : $\lambda_G$ : $N_{IE}$}
		& Prec. $\uparrow$ & $\Delta_{Prec.}$
		& AUC $\uparrow$& $\Delta_{AUC}$ \\
		\midrule
		1 & 2.0 : 5.0 : 0.0 & 58.8 & -4.5 & 47.2 & -2.4 \\
		2 & 2.0 : 5.0 : 1.0 & 59.3 & -4.0 & 47.3 & -2.3 \\
		\textbf{3} & \textbf{2.0 : 5.0 : 2.0} & \textbf{63.3} & \textbf{--} & \textbf{49.6} & \textbf{--} \\
		4 & 2.0 : 5.0 : 3.0 & 63.0 & -0.3 & 48.8 & -0.8 \\
		\midrule  
		5 & 1.0 : 3.5 : 2.0 & 58.2 & -5.1 & 46.7 & -2.9 \\
		6 & 1.0 : 4.5 : 2.0 & 58.9 & -4.4 & 46.1 & -3.5 \\
		7 & 2.0 : 5.5 : 2.0 & 61.5 & -1.8 & 48.3 & -1.3 \\
		8 & 3.0 : 5.5 : 2.0 & 61.0 & -2.3 & 48.1 & -1.5 \\
		\midrule
		- & Dataset Ratio & - & - & - & - \\
		\midrule
		9 & 1 : 1 : 1 : 1 & 60.9 & -2.4 & 46.9 & -2.7 \\
		10 & 1 : 1 : 2 : 2 & 62.3 & -1.0 & 49.1 & -0.5 \\
		11 & 2 : 1 : 1 : 2 & 61.4 & -1.9 & 48.2 & -1.4 \\
		12 & 2 : 2 : 1 : 1 & 60.3 & -2.0 & 47.1 & -2.5 \\
		\textbf{13} & \textbf{2 : 2 : 3 : 2} & \textbf{63.3} & \textbf{-} & \textbf{49.6} & \textbf{-} \\
		\bottomrule
	\end{tabularx}
	\label{tab:para_ablation}
\end{table}

\begin{table}[h]
	\caption{Ablation study in terms of Params, FLOPs, and Speed on NAT2021-L.}
	\small
	\centering
	\renewcommand{\arraystretch}{1.1}
	\begin{tabularx}{\linewidth}{c|l *{5}{>{\centering\arraybackslash}X}}
		\toprule
		\# & Trackers & Params & FLOPs & FPS & 
		\multicolumn{1}{c}{Prec.$\uparrow$} & 
		\multicolumn{1}{c}{AUC $\uparrow$} \\ 
		\midrule
		1 & Baseline     & 93M  & 29G  & 60 & 59.9 & 47.7 \\
		2 & +DPP         & 94M  & 29G  & 58 & 60.7 & 47.9 \\
		3 & +DPP+DKE     & 103M & 34G  & 49 & 62.4 & 48.8 \\
		4 & +DKE+KGP     & 103M & 34G  & 49 & 63.1 & 49.1 \\
		\textbf{5} & \textbf{DPTrack} & \textbf{104M} & \textbf{34G} & \textbf{49} & \textbf{63.3} & \textbf{49.6} \\
		\bottomrule
	\end{tabularx}
	\label{tab:param}
\end{table}

\subsubsection{Ablation Study of Param-Scale}We ablate parameter scale (Table \ref{tab:param}) of each component. Compared with the baseline (\textbf{93M} params, \textbf{29G} FLOPs), DPTrack adds only minor overhead (\textbf{104M}, \textbf{34G}), while still running in real time (\textbf{49 FPS}).
Concretely, adding DPP increases the model by only \textbf{1M} parameters with no extra FLOPs (\textbf{93M/29G} to \textbf{94M/29G}), causing a small FPS drop (\textbf{60} to \textbf{58}) while improving accuracy. 
Incorporating DKE raises the complexity to \textbf{103M/34G} and lowers the throughput to 49 FPS, yet delivers larger gains than DPP. Replacing DPP with KGP at the same complexity (103M/34G, 49 FPS) yields further accuracy gains.
The full DPTrack then adds only \textbf{1M} parameters and achieves the best performance (\textbf{63.3\%} precision, \textbf{49.6\%} AUC). 
It is worth noting that the variant without DKE is not included, DKE as the bridge facilitating interaction between KGP and DPP, thereby enabling the tracker to exhibit strong robustness against glare. In summary, DPTrack achieves significant performance gains with only marginal increases in parameters and computation, highlighting the efficiency of its modular design.
\begin{figure*}[t]
	\centering
	\includegraphics[width=0.99\textwidth]{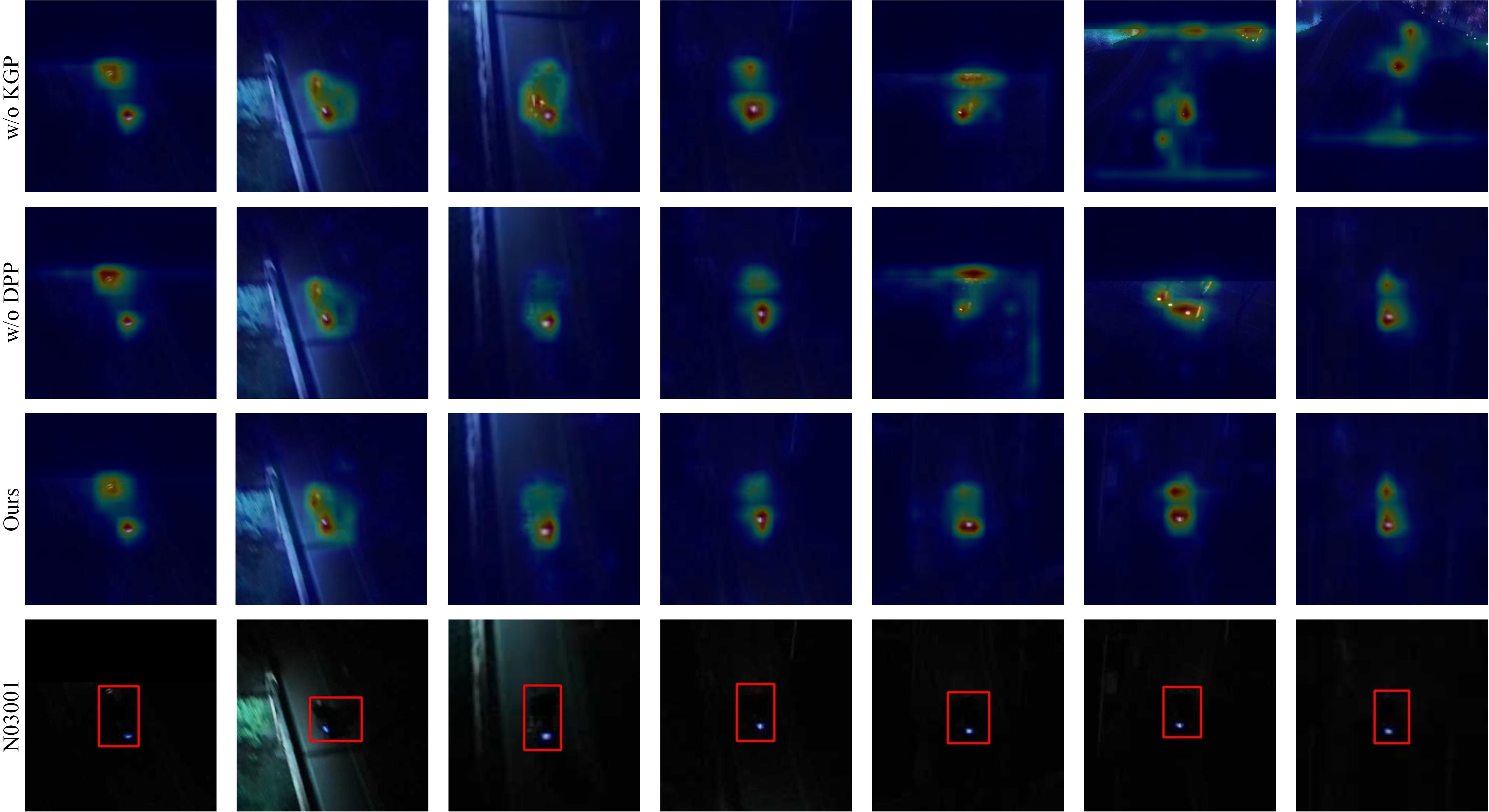}
	\caption{We visualize the results of the ablation study on key modules in the N03001 sequence under nighttime scenarios. The regions of interest captured by different models at the same frame are extracted, and the corresponding heatmaps directly reveal the contribution of each module.}
	\label{fig:vis_abli_nat2021l}
\end{figure*}

\begin{figure*}[t]
	\centering
	\includegraphics[width=0.99\textwidth]{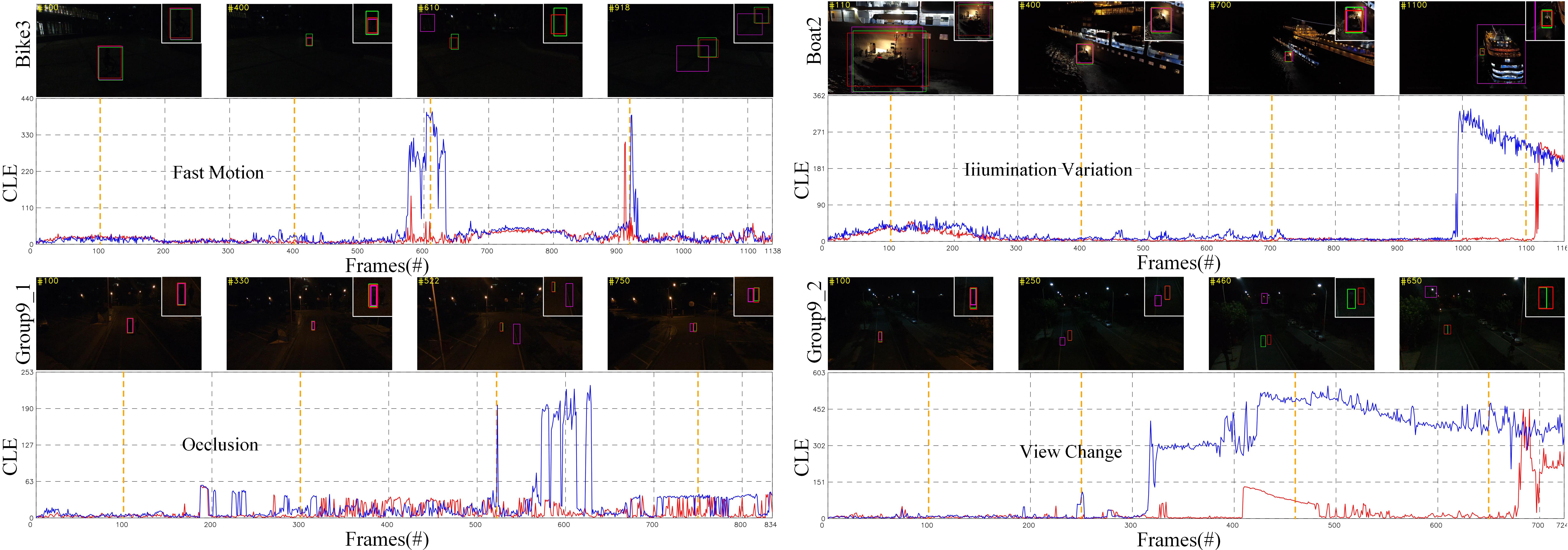} 
	\caption{Generalization evaluation under challenging nighttime scenarios, where CLE curves of DPTrack (red) and ORTrack \cite{c:ortrack} (blue) are compared under fast motion, illumination variation, occlusion, and view change. DPTrack exhibits lower errors and stronger robustness across different challenges.}
	\label{fig:CLE}
\end{figure*}

\subsubsection{Ablation-Based Visualization} To elucidate the roles of each component, we visualize local heatmaps from DPTrack and its variants (Fig.~\ref{fig:vis_abli_nat2021l}). Without KGP, the tracker lacks precise guidance and distinguishes candidates only coarsely, producing diffuse responses and weakened object focus. Ablating DPP prevents modeling global–local topology, the tracker perceives only salient local cues, fails to integrate them holistically, and shows dispersed activations with incomplete localization. By contrast, DPTrack integrates global-to-local representations and produces compact, accurate heatmaps, demonstrating robustness in nighttime scenes. DKE is retained because it bridges KGP and DPP, by mitigating local glare, it substantiates KGP–DPP complementarity and the necessity of DKE for robust nighttime aerial tracking.
\subsection{Analysis on Generalization Evaluation}
We evaluate DPTrack’s generalization using center location error (CLE) on four UAVDark135 sequences with diverse attributes. The red and blue curves denote DPTrack and ORTrack\cite{c:ortrack}, respectively, with representative frames shown at key points.
\subsubsection{Fast Motion}
In the Bike3 sequence characterized by fast motion, the CLE curves reveal that DPTrack maintains consistently low errors across most frames, whereas ORTrack exhibits large error spikes, particularly around frame \#610. Under dark conditions with rapid object movement, ORTrack fails to effectively discriminate target features from background noise and drifts significantly, while DPTrack successfully keeps the object centered. This demonstrates that DPTrack can effectively handle motion-induced feature blurring and exhibits strong generalization capability.
\subsubsection{Illumination Variation}
In the Boat2 sequence characterized by illumination variation, the CLE curves show that DPTrack maintains low errors throughout, while ORTrack exhibits noticeable fluctuations and a sharp error increase after frame \#1000. As illumination changes drastically across frames, ORTrack struggles to adapt to brightness variation and background interference, often misinterpreting bright regions as part of the target. Nevertheless, when illumination is relatively uniform (e.g., frame \#110), it can still estimate the target scale accurately. In contrast, DPTrack remains stable and precisely localizes the object, indicating stronger robustness to illumination variation.
\subsubsection{Occlusion}
Occlusion is common in crowded scenarios. The consistently low error curves indicate that DPTrack discriminates candidate features more precisely, while ORTrack suffers frequent error spikes, especially between frames \#500–\#700 under severe occlusion. When the object is partially or fully blocked, ORTrack often fails to re-identify it and drifts toward salient background regions. In contrast, DPTrack extracts robust features for fine-grained localization after occlusion, maintaining stable tracking. These results demonstrate DPTrack’s strong adaptability to nighttime scenarios with heavy occlusion.
\subsubsection{View Change}
Nighttime aerial tracking often involves dynamic viewpoints, where feature transformations pose severe challenges for trackers. As the viewpoint changes, object appearance varies accordingly, and the CLE curves indicate that ORTrack suffers persistent large errors, especially after frame \#400 when a major viewpoint shift occurs, completely losing the target. In contrast, DPTrack maintains low errors across most frames. Although it still struggles under drastic appearance changes, once the object stabilizes, it quickly re-locks onto the target, demonstrating strong generalization under dynamic viewpoint variations in nighttime scenes.

\section{Limitation and Future Work}
Although the proposed aerial tracker can accurately track ground target under nighttime conditions, certain shortcomings persist. First, DPTrack mainly focuses on low-illumination challenges in nighttime scenarios, while its robustness under other adverse weather conditions (e.g., rain and fog) has not been investigated. Second, the tracker relies on a pre-trained backbone for similarity matching, which cannot be jointly optimized with the prompt generation pipeline owing to dataset scale discrepancies. This non-end-to-end optimization may limit inter-module collaboration and compromise the overall consistency of feature learning. In future work, we will investigate cross-scene generalization based on a more comprehensive dataset to address the above limitations, with the ultimate goal of developing a robust aerial tracking system capable of operating under complex environmental conditions.

\section{Conclusion}
In this paper, we present DPTrack, a novel nighttime aerial tracker that leverages fine-grained guidance signals derived from intrinsic attributes to generate precise prompts, thereby enabling accurate object localization. Specifically, we propose a prompt optimization strategy built upon fine-grained cues extracted from the object’s intrinsic attributes. The strategy integrates three complementary modules: a topology-aware DPP, a feature-selectivity guided DKE, and a class–channel correspondence based KGP, which collaboratively produce location-specific prompts to guide precise tracking. Extensive experiments demonstrate that DPTrack achieves superior object perception while maintaining robust tracking performance.

\bibliographystyle{IEEEtran}
\bibliography{IEEEabrv,DPTrack}

\end{document}